\documentclass[10pt,twocolumn,letterpaper]{article}

\usepackage[pagenumbers]{cvpr} %

\usepackage[dvipsnames]{xcolor}

\makeatletter
\def\adl@drawiv#1#2#3{%
        \hskip.5\tabcolsep
        \xleaders#3{#2.5\@tempdimb #1{1}#2.5\@tempdimb}%
                #2\z@ plus1fil minus1fil\relax
        \hskip.5\tabcolsep}
\newcommand{\cdashlinelr}[1]{%
  \noalign{\vskip\aboverulesep
           \global\let\@dashdrawstore\adl@draw
           \global\let\adl@draw\adl@drawiv}
  \cdashline{#1}
  \noalign{\global\let\adl@draw\@dashdrawstore
           \vskip\belowrulesep}}
\makeatother

\usepackage{times}
\usepackage{epsfig}
\usepackage{graphicx}
\usepackage{amsmath}
\usepackage{amssymb}
\usepackage{booktabs}
\usepackage{multirow}
\usepackage{xcolor}
\usepackage{arydshln}
\usepackage{wrapfig}
\usepackage{enumitem}
\usepackage[font=small,hypcap=false]{caption}
\usepackage{tabularx} 
\usepackage{microtype}
\usepackage{pifont}
\usepackage{listings}
\usepackage{cuted}
\usepackage{capt-of} %

\definecolor{codegreen}{rgb}{0,0.6,0}
\definecolor{codegray}{rgb}{0.5,0.5,0.5}
\definecolor{codepurple}{rgb}{0.58,0,0.82}
\definecolor{backcolour}{rgb}{0.95,0.95,0.92}

\lstdefinestyle{mystyle}{
    backgroundcolor=\color{backcolour},   
    commentstyle=\color{codegreen},
    keywordstyle=\color{magenta},
    numberstyle=\tiny\color{codegray},
    stringstyle=\color{codepurple},
    basicstyle=\ttfamily\footnotesize,
    breakatwhitespace=false,         
    breaklines=true,                 
    captionpos=b,                    
    keepspaces=true,                 
    numbers=left,                    
    numbersep=5pt,                  
    showspaces=false,                
    showstringspaces=false,
    showtabs=false,                  
    tabsize=2,
    showlines=true
}

\newcommand{\cmark}{\ding{51}}

\newcommand\rgt{\aftergroup\mathclose\aftergroup{\aftergroup}\right}
\newcommand{\ci}[1]{\scriptsize{\textcolor{gray}{~($\pm #1$)}}}

\newcommand{\supparxiv}[2]{#2}

\supparxiv{

\def\upvspacefig{\vspace{-6mm}}
\def\downvspacefig{\vspace{-4mm}}
\setlength{\belowcaptionskip}{-0mm}
\setlength{\abovecaptionskip}{0.3em}

}{
\def\upvspacefig{\vspace{-0mm}}
\def\downvspacefig{\vspace{-0mm}}
\setlength{\belowcaptionskip}{-0mm}
\setlength{\abovecaptionskip}{0.5em} 

}

\makeatletter
\renewcommand\paragraph{\@startsection{paragraph}{4}{\z@}%
	{0.7ex \@plus.3ex \@minus.2ex}%
	{-1em}%
	{\normalfont\normalsize\bfseries\maybe@addperiod}}
\newcommand{\maybe@addperiod}[1]{#1\@addpunct{.}}
\makeatother

\def\projecturl{https://ificl.github.io/MultiFoley/}

\definecolor{cvprblue}{rgb}{0.21,0.49,0.74}
\definecolor{urlblue}{rgb}{0.24,0.49,0.9}
\usepackage[pagebackref,breaklinks,colorlinks,urlcolor=urlblue,citecolor=cvprblue]{hyperref}

\def\paperID{1649} %
\def\confName{CVPR}
\def\confYear{2025}

\title{Video-Guided Foley Sound Generation with Multimodal Controls}

\author{Ziyang Chen\textsuperscript{1,2$^{*}$}\qquad 
Prem Seetharaman\textsuperscript{2}\qquad 
Bryan Russell\textsuperscript{2} \qquad
Oriol Nieto\textsuperscript{2} \vspace{1.2mm}\\
David Bourgin\textsuperscript{2} \qquad 
Andrew Owens\textsuperscript{1} \qquad 
Justin Salamon\textsuperscript{2}
\vspace{2mm}\\
\textsuperscript{1}University of Michigan\quad \textsuperscript{2}Adobe Research \vspace{1mm}\\
{\normalsize \texttt{\url{\projecturl}}}
}

\begin{document}
\maketitle

{\let\thefootnote\relax\footnotetext{{* Work done during an internship at Adobe.}}}

\begin{strip}
\centering
\supparxiv{\vspace{-11mm}}{\vspace{-6mm}}

\begin{flushleft}
    \vspace{-10mm}
    \hspace{3em} (a) Foley with text control 
    \hspace{5em} (b) Foley with audio control
    \hspace{5em} (c) Foley audio extension
    \vspace{-3mm}
\end{flushleft}   
    \includegraphics[width=\textwidth]{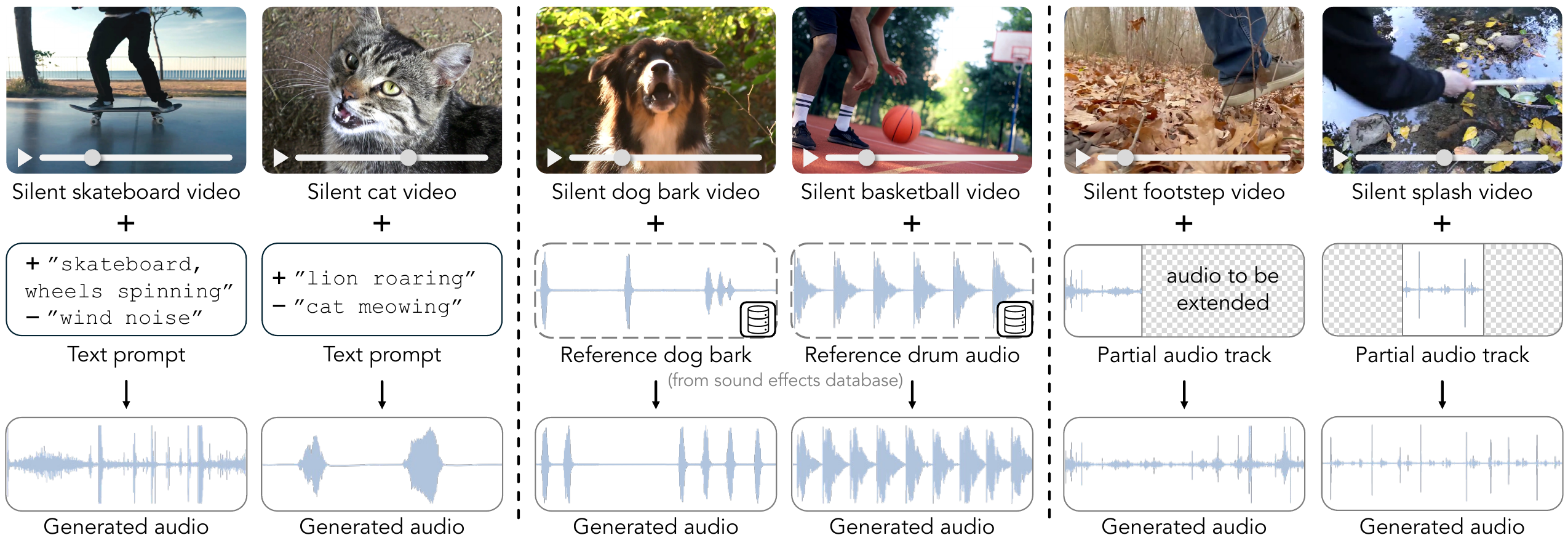}

\captionof{figure}{\textbf{MultiFoley for video-guided sound generation with multimodal controls.} We generate Foley sounds for silent videos with various control signals to shape their audio. (a) Text prompts, both positive and negative, guide synchronized Foley generation. (b) Reference audio from sound libraries defines the customized audio style. (c) A partial audio track is extended to produce a complete Foley sound. {\bf We encourage the reader to watch and listen to the results in our \href{\projecturl}{website}.} }

\label{fig:teaser}
\end{strip}

\begin{abstract}
Generating sound effects for videos often requires creating artistic sound effects that diverge significantly from real-life sources and flexible control in the sound design. 
To address this problem, we introduce {\em MultiFoley}, a model designed for video-guided sound generation that supports multimodal conditioning through text, audio, and video. 
Given a silent video and a text prompt, MultiFoley allows users to create clean sounds (\eg, skateboard wheels spinning without wind noise) or more whimsical sounds (\eg, making a lion's roar sound like a cat's meow).
MultiFoley also allows users to choose reference audio from sound effects (SFX) libraries or partial videos for conditioning. 
A key novelty of our model lies in its joint training on both internet video datasets with low-quality audio and professional SFX recordings, enabling high-quality, full-bandwidth (48kHz) audio generation.
Through automated evaluations and human studies, we demonstrate that MultiFoley successfully generates synchronized high-quality sounds across varied conditional inputs and outperforms existing methods. Please see our project page for video results: \small \texttt{\url{\projecturl}}

\end{abstract}

\supparxiv{\vspace{-3mm}}

\section{Introduction}
\label{sec:intro}

Sound designers, often create sound effects from different sources that may not resemble the original video, such as everyday objects or pre-recorded sound libraries, to create a soundtrack for a video that
synchronizes with on-screen actions. For example, they use crinkling paper for a warm fire, a coconut shell for horse hooves, or the cracking of celery for breaking bones. This process is known as {\em Foley}~\cite{ament2014foley}. Instead of replicating the true sounds, sound designers aim to achieve an artistic effect that enhances the viewers' experience.

Recent works~\cite{mei2023foleygen,luo2024diff,wang2024frieren, wang2024v2a,su2024vision} mainly formulated the Foley problem as a video-to-audio generation task, generating sound effects that align temporally and semantically with the video.
However, those approaches restrict audio content and lack the control designers need, as they often create sounds that diverge from real life in many different ways. 
Current systems only offer limited controls (\eg, conditional video examples~\cite{du2023conditional} or language~\cite{xie2024sonicvisionlm}) and face issues with audio quality and synchronization.

To give users more creative control over sound design, we propose \textsc{MultiFoley}, a video-guided Foley sound generation framework that integrates multimodal controls, using text, audio, and video conditional signals.
Our model provides sound designers fine-grained control over how the audio sounds while easing the burden of synchronizing audio to videos. 
By jointly training across audio, video, and text modalities, \textsc{MultiFoley} enables flexible control and various Foley applications.
Users can customize audio content through text prompts -- whether or not they match the visuals -- while maintaining synchronization. For example, users can generate clean sound effects by removing unwanted elements like wind noise using text prompts or they can replace a cat meow with a lion roar, as shown in \cref{fig:teaser}. Beyond text, our system can also accept different types of audio and audio-visual conditioning. Users can generate desired sound effects from reference audio in the sound effects (SFX) library or extend the soundtrack from a portion of a video. 

One of the key challenges is that internet video soundtracks are often poorly aligned with the visual content (\eg, irrelevant audio) and suffer from low quality (\eg, noisy audio and limited bandwidth).
To address this, we jointly train on high-quality SFX libraries (audio-text pairs) alongside internet videos, using language supervision in both cases. This approach enables our model to generate full-bandwidth audio (48kHz) that meets professional standards and enhances precise text-based customization.

\textsc{MultiFoley} consists of a diffusion transformer, a high-quality audio autoencoder (based on \cite{kumar2024high}), a frozen video encoder for audio-video synchronization, and a novel multi-conditional training strategy enabling flexible downstream tasks like audio extension and text-driven sound design.
Our key contributions are as follows:

\begin{itemize}[leftmargin=10pt, topsep=1pt, noitemsep]
\item We present a unified framework for video-guided Foley generation that leverages multiple conditioning modalities—text, audio, and video—within a single model.
\item We introduce a training approach that combines internet video datasets with low-quality audio and professional libraries via language as bridges to enable high-quality, full-bandwidth audio generation at 48kHz.
\item We show that our model supports a diverse range of applications, such as text-controlled Foley generation, audio-controlled Foley generation, and Foley extension, expanding possibilities for creative audio production.
\item Through extensive quantitative evaluations and human studies, we demonstrate that MultiFoley achieves better audio quality and improved cross-modal alignment, outperforming existing methods in key benchmarks.
\end{itemize}

\section{Related Work}
\label{sec:related_work}

\paragraph{Video-to-audio generation}
Video-to-audio generation has recently gained significant attention. 
Several works have employed auto-regressive transformer models to generate audio from visual features, \eg, SpecVQGAN~\cite{iashin2021taming}, Im2Wav~\cite{sheffer2023hear}, FoleyGen~\cite{mei2023foleygen}, and V-AURA~\cite{viertola2024temporally}. 
Other approaches have introduced video-to-audio diffusion or flow matching models to address this task, \eg, Diff-Foley~\cite{luo2024diff}, Action2Sound~\cite{chen2024action2sound}, VTA-LDM~\cite{xu2024video}, and Frieren~\cite{wang2024frieren}. 
Some researchers have adapted the MaskGIT~\cite{chang2022maskgit} framework for video-to-audio synthesis~\cite{pascual2024masked, su2024vision,liu2024tellhearvideo}. 
Other approaches use a two-stage process: first, extracting time-varying signals like sound onsets or energy curves from videos, then adapting pretrained text-to-audio diffusion models with specialized adapters for video-to-audio generation with optional text~\cite{xie2024sonicvisionlm, comunita2024syncfusion,zhang2024foleycrafter, jeong2024read, huang2024rhythmic, lee2024video}. V2A-Mapper~\cite{wang2024v2a} translates CLIP visual embeddings into CLAP space to condition audio generation via AudioLDM~\cite{liu2023audioldm}.
Recently, {\em concurrent unpublished} work Movie Gen Audio~\cite{polyak2024movie} explores generating audio conditioned on video and text. This work demonstrates that text provides complementary information for audio generation. Concurrent work MMAudio~\cite{cheng2024taming} also explores multimodal joint training by combining video-to-audio and text-to-audio generation across multiple datasets.
In contrast, our approach stands out by providing richer multimodal controls and Foley applications, such as generating on-screen sounds with semantically different audio via text and audio-conditional Foley generation, as shown in \cref{fig:teaser}.

\paragraph{Diffusion models}
Diffusion models~\cite{sohldickstein2015diffusion,ho2020denoising,song2020score,dhariwal2021diffusion,song2020denoising} are generative models that learn to reconstruct data by reversing a process in which data is gradually corrupted with noise. By iterative denoising, these models generate new samples starting from random noise.
Latent diffusion models~(LDMs)~\cite{rombach2022high} perform the diffusion process in a latent space by translating data using a pretrained encoder and decoder pair. These models have proven useful for various generative applications, such as image generation and editing~\cite{dhariwal2021diffusion,rombach2022high,nichol2021glide,saharia2022imagen,meng2021sdedit,hertz2022prompt,brooks2023instructpix2pix, geng2024motion}, video generation~\cite{ho2022imagen,singer2022make,bar2024lumiere,girdhar2023emu}, audio generation~\cite{liu2023audioldm2,evans2024fast,evans2024stable,majumder2024tango,xue2024auffusion, li2025self}, and 3D generation~\cite{hoellein2023text2room,liu2023zero,bensadoun2024meta}. They have also been applied to tasks such as semantic segmentation~\cite{amit2021segdiff, xu2023open}, camera pose estimation~\cite{wang2023posediffusion, zhang2024raydiffusion}, depth estimation~\cite{saxena2023monocular,ke2024repurposing}, and compositional generation~\cite{liu2022compositional,bar2023multidiffusion,geng2023visual,geng2024factorized,chen2024images}. 
In our work, we leverage diffusion models for video-guided Foley sound generation, incorporating controls from multiple modalities.

\paragraph{Audio-visual learning}
Many works have focused on learning multimodal representations from audio-visual data. Some study {\em semantic correspondence} between sight and sound, \eg, learning common audio-visual associations~\cite{arandjelovic2017look,morgado2021audio,girdhar2023imagebind,asano2020labelling,lin2024siamese}, audio-visual sound localization~\cite{arandjelovic2018objects,hu2022mix,park2024can}, and audio-visual segmentation~\cite{zhou2023audio,liu2024audio}. Some investigate \textit{temporal correspondence} between audio and video~\cite{owens2018audio,chen2021audio,feng2023self,sun2023eventfulness,iashin2024synchformer,nugroho2023audio}, studying how sounds synchronize with visual events over time. Some explore \textit{spatial correspondence} between audio and visual data, particularly how sound conveys information about the spatial environment~\cite{chen2021structure,gao2020visualechoes,yang2020telling,chen2022sound,chen2023sound,majumder2023learning,chen2023everywhere,chen2024RAF}. Inspired by them, we propose a joint learning approach across text, audio, and video modalities, enabling flexible control over various aspects of generated audio.

\begin{figure}[!t]
    \centering
    \upvspacefig
    \includegraphics[width=0.9\linewidth]{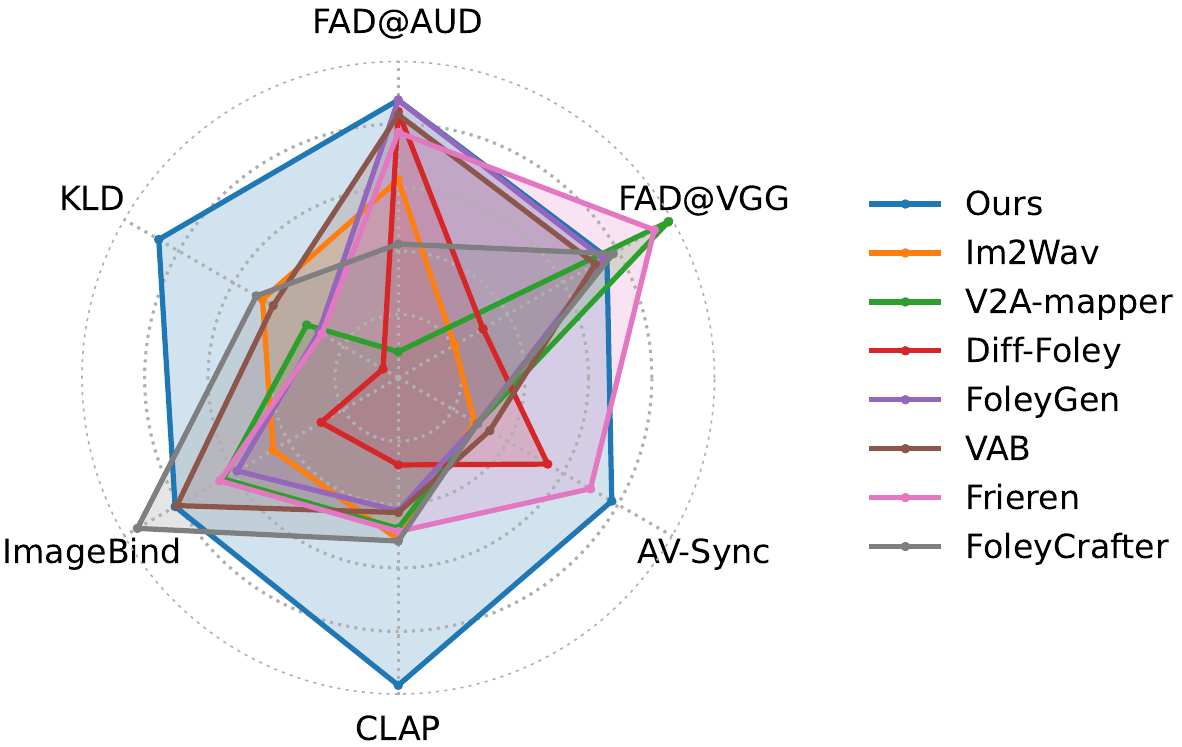}
    
    \caption{\textbf{Radar chart comparison for video-to-audio generation task.} Each metric is normalized for a better visualization. 
    } 
    \downvspacefig
    \label{fig:video2audio}
\end{figure}

\paragraph{Generative models for sound design} 
Foley sound design has traditionally relied on Foley artists to create synchronized sound effects using everyday objects or sound libraries~\cite{ament2014foley}. Recent text-to-audio models~\cite{evans2024fast,evans2024stable,majumder2024tango} give sound designers flexible control, allowing them to generate audio attributes directly from descriptive text. However, sound designers still need to manually synchronize the generated audio with the video.
T-Foley~\cite{chung2024t} proposes to address this issue by guiding foley generation using temporal event features like the RMS of the waveform.
Video-to-audio models~\cite{wang2024frieren,viertola2024temporally} generate synchronized audio but lack user control options. \citet{du2023conditional} introduces a conditional Foley task for generating soundtracks from additional video examples. 
Building on the work above, our model offers artists improved tools for generating synchronized audio with versatile controls using text, audio, and video.

\begin{figure*}[t]
\upvspacefig
\centering

\includegraphics[width=1.0\linewidth]{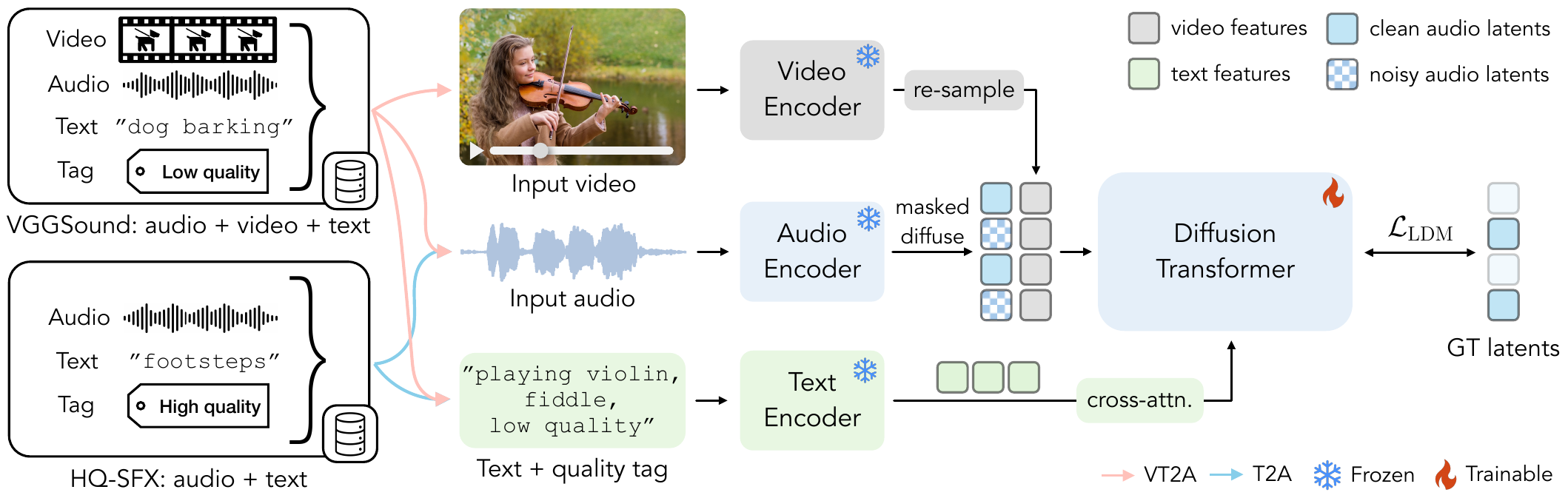}

\caption{ {\bf Method overview.} 
We train our model jointly on a standard audio-video dataset {\em VGGSound} for VT2A generation and a high-quality audio-text dataset {\em HQ-SFX} for T2A generation.
We encode the input audio into latents, adding noise to a portion of them. The silent video is encoded into visual features, concatenated with the audio latents along the channel dimension. The text input, including a quality tag, is encoded through a text encoder and applied via cross-attention. }

\downvspacefig
\label{fig:method}
\end{figure*}

\section{Multimodal Conditional Foley Generation}
\label{sec:method}

\def\bt{{\mathbf t}} %
\def\ba{{\mathbf a}} %
\def\bv{{\mathbf v}} %
\def\bz{{\mathbf z}} %

\def\dm{\boldsymbol{\epsilon}_{\theta}} %

We aim to generate synchronized Foley sounds for silent videos, guided by multimodal inputs like text, audio, or video examples that define how the video should sound. First, we introduce a simple framework called \textsc{MultiFoley} that performs video-guided Foley generation with multimodal conditions, shown in \cref{fig:method}. Then, we show various applications our model enables for conditional Foley generation.

\subsection{Generative Formulation} %
Given an input silent video $\bv_q$, our goal is to generate a synchronized soundtrack $\hat{\ba}$. This generation is conditioned on a text prompt $\bt_c$ and a reference audio-visual pair ($\ba_c$, $\bv_c$), which can be part of $\bv_q$ or from different videos.
We frame this problem as a conditional diffusion task. We learn a diffusion model $\dm$, parameterized by $\theta$, to model the conditional distribution $p_{\theta}(\ba | \bv_q, \bt_c, \ba_c, \bv_c)$ from the audio $\ba$.  

We perform the diffusion process in a latent space~\cite{rombach2022high}. We use a pretrained audio latent encoder $\mathcal{E}_a$ and decoder $\mathcal{D}_a$ that converts the waveform $\ba \in \mathbb{R}^{T_a}$ into the compressed latent code $\bz = \mathcal{E}_a(\ba) \in \mathbb{R}^{T_z \times C_z}$, where $T_a$ and $T_z$ represent the lengths of the waveform and audio latents respectively, and $C_z$ denotes the dimension of the latent features. 
During the forward diffusion process, we gradually add Gaussian noise to clean audio latents $\bz_0$ to obtain noisy latent $\bz_t$ at different timesteps $t \in \{ 1, ..., T\}$. During the reverse process, the denoiser $\dm(\cdot)$ computes the noise estimates $\hat{\boldsymbol{\epsilon}} = \dm(\bz_t, \bv_q, \bt_c, \ba_c, \bv_c, t)$ and optimize following \cite{ho2020denoising,rombach2022high}: 
\begin{equation}
\mathcal{L}_{\mathrm{LDM}} = \mathbb{E}_{\boldsymbol{\epsilon} \sim \mathcal{N}(0, \mathbf{I}), t \sim \mathcal{U}(T)} \Big[ \Vert \boldsymbol{\epsilon} - \hat{\boldsymbol{\epsilon}}
\Vert_{2}^{2}\Big]\text{,}
\end{equation}
where conditions are encoded by the visual encoder $\mathcal{E}_v$, audio encoder $\mathcal{E}_a$ or text encoder $\mathcal{E}_t$, respectively. 
During inference, we iteratively denoise the Gaussian noise $\bz_T$ with the noise estimates $\hat{\boldsymbol{\epsilon}}$ to obtain the final clean latent $\hat{\bz}_0$, which is then decoded into the waveform $\hat{\ba} =\mathcal{D}_a(\hat{\bz}_0)$.

\subsection{Multimodal Conditioning and Training}
\label{sec:model-recipe}
We build our model on the Diffusion Transformer (DiT)~\cite{peebles2023scalable} and a pretrained DAC-VAE~\cite{kumar2024high} for the audio encoder and decoder, as shown in \cref{fig:method}.

\paragraph{Multimodal conditioning}

We use a pretrained visual encoder $\mathcal{E}_v$ to encode the silent video $\bv_q \in \mathbb{R}^{T_v \times 3 \times H \times W}$ into features $\mathcal{E}_v(\bv_q) \in \mathbb{R}^{T_v \times C_v}$, where $T_v$ is the number of video frames, $H, W$ are video height and width, and $C_v$ is the visual feature dimension size. These video features are interpolated to match the length of the audio latents $\bz$, then concatenated along the channel dimension with noisy audio latents $\bz_t$ to ensure the audio-visual temporal alignment, pairing audio and video features at the same time, similar to Wang \etal~\cite{wang2024frieren}. The combined features are passed to the feed-forward transformer $\dm(\cdot)$. 

For the text condition $\bt_c$, we apply a frozen text encoder $\mathcal{E}_t$ to extract embeddings $\mathcal{E}_t(\bt_c) \in \mathbb{R}^{T_t \times C_t}$, where $T_t$ is the tokenized text sequence length, and $C_t$ is token’s embedding dimension. These text embeddings are then incorporated into $\dm(\cdot)$ via cross-attention.

To improve our model's ability to condition on audio and video, during training we provide the model with a sample of ground truth audio-visual segments,  \(\mathcal{E}_a(\ba_c)\) and \(\mathcal{E}_v(\bv_c)\), which we select randomly from the clean video. %
Our training loss is then only applied to the latents that were not provided as a conditioning signal.  At inference, we can obtain our audio-visual conditioning, \((\ba_c, \bv_c)\), from a {\em different} video.

\paragraph{Learning from combined datasets}
To enhance audio generation quality and text control, we train our model jointly on an audio-visual(-text) dataset, VGGSound~\cite{chen2020vggsound}, and a large proprietary licensed dataset of high-quality sound effects with text captions, termed as {\em HQ-SFX}.
The audio in these datasets is quite different, since VGGSound is ``in-the-wild'' and relatively noisy, whereas HQ-SFX contains professionally recorded sound effects and thus is generally higher quality and full bandwidth.
To give the model control over audio quality, we assign {\em quality tags} to both datasets, which represent different dataset distributions: VGGSound examples are paired with a ``low quality'' tag in prompts, \eg, ``{\tt \small {category}, low quality}''. In contrast, HQ-SFX examples are labeled with a ``high quality'' tag in their captions. %
During training, we randomly drop out the caption or quality tag to allow the model to disentangle audio quality from semantics and encourage the model to learn and distinguish audio quality across different data distributions.

\paragraph{Implementation}
Our denoiser $\dm$ consists of a 12-layer DiT. We train a DAC-VAE to map 48kHz audio waveforms into 40Hz latent features with a channel dimension $C_z$ of 64, the encoder is frozen during the training. We use the T5-base text encoder~\cite{raffel2020exploring} to obtain embeddings with $C_t=768$, and we apply CAVP~\cite{luo2024diff} to extract visual features at 8 FPS for 8-sec videos with $T_v=64$ and $C_v=512$. 

The model is jointly trained for video-text-to-audio generation on VGGSound, containing 168K samples, and for text-to-audio generation on the HQ-SFX dataset, with 400K samples. We balance the datasets to ensure robust training and train the model for 600K iterations with a batch size of 128. During training, audio latents are masked with a probability of 0.25, where the masking spans 0 to 2 seconds. We also randomly drop the video, caption, and quality tag conditions during the training. 
We then finetune this pretrained model for 50k iterations on a curated subset of VGGSound, selected for high audio-visual correspondence using an ImageBind~\cite{girdhar2023imagebind} score threshold of 0.3.
During inference, we use the DDIM~\cite{song2020denoising} method and
apply a classifier-free guidance scale ranging from 3.0 to 5.0 to denoise over 100 steps. \supparxiv{Please see more details in the supplementary material.}{See \cref{appendix:implement} for more details.}

\begin{table*}[!t]
\upvspacefig

\caption{{\bf Evaluation of video-to-audio generation.} Each method is evaluated on the VGGSound test set across six metrics assessing cross-modal alignment and audio quality. ImageBind and CLAP scores are reported in \%. The unit of AV-Sync is seconds. \textcolor{gray}{DAC-VAE} reconstructs the VGGSound audio and serves as an oracle baseline. The best results are in {\bf bold}.}
\label{tab:video2audio}

\centering
\footnotesize
\setlength\tabcolsep{7pt}
\renewcommand{\arraystretch}{1.0}
\newcommand\padd{\phantom{0}}

\resizebox{1\textwidth}{!}{

\begin{tabular}{lc ccc ccc}
\toprule

\multirow{3}{*}{Method}  & \multirow{3}{*}{\shortstack[c]{Sampling\\rate~(Hz)}} & \multicolumn{3}{c}{Cross-modal alignment}   & \multicolumn{3}{c}{Audio quality}   \\
\cmidrule(lr){3-5} \cmidrule(lr){6-8} 
 &  &  ImageBind~$\uparrow$ & CLAP~$\uparrow$ & AV-Sync~$\downarrow$ & FAD@VGG~$\downarrow$ & FAD@AUD~$\downarrow$ & KLD~$\downarrow$     \\
\midrule
Im2Wav~\cite{sheffer2023hear} & 16K  &  22.3 &  25.2 & 1.25 & 7.36 & 5.88 & 2.11 \\
V2A-mapper~\cite{wang2024v2a} & 16K  &  25.2 & 24.5 & 1.24 & \textbf{1.13}  & 8.59  &   2.40 \\
Diff-Foley~\cite{luo2024diff} &  16K  & 19.5 &  20.5 & 1.01  &  6.53 &  4.80 & 2.90 \\
FoleyGen~\cite{mei2023foleygen} & 16K & 24.4 & 23.4 & 1.24 & 3.01 & \textbf{4.62} & 2.48  \\
VAB~\cite{su2024vision} & 16K  & 27.9   & 23.5 & 1.20 &  3.26 &  4.85 & 2.18\\
Frieren~\cite{wang2024frieren} &  16K & 25.4  & 24.7 & 0.87 & 1.54  & 5.13  &   2.50 \\
FoleyCrafter~\cite{zhang2024foleycrafter} &  16K  &  \textbf{ 30.2 }  & 25.3 & 1.24 &  2.74 & 6.89 & 2.07\\
\cdashlinelr{1-8}

MultiFoley (ours)  & \textbf{48K}  &  28.0   &  \textbf{34.4} & \textbf{0.80} & 2.92 & \textbf{4.62}  & \textbf{1.43} \\

\midrule
\textcolor{gray}{DAC-VAE~\tiny{(VGGSound)}} & \textcolor{gray}{48K}  & \textcolor{gray}{35.4} & \textcolor{gray}{28.2} & \textcolor{gray}{0.62}  & \textcolor{gray}{1.21} & \textcolor{gray}{5.91}  & \textcolor{gray}{0.28} \\

\bottomrule
\end{tabular}
\downvspacefig
}
\end{table*}

\subsection{Foley Applications}

\paragraph{Video-guided Foley with text control}
Training audio diffusion with both video and text inputs allows our model to generate Foley sounds controlled by text. This dual conditioning setup enables flexible audio generation, where users can influence sound characteristics based on text prompts.

Our approach associates language with video cues, disentangling the semantic and temporal elements of videos. This setup allows for creative Foley applications, such as modifying a bird-chirping video to sound like a human voice or transforming a typewriter’s sound into piano notes -- all while remaining synchronized with the video. Also, negative prompting provides a way to exclude unwanted audio elements by specifying them in the negative text prompts, offering flexible control over the audio output.
To achieve this application, we use classifier-free guidance (CFG)~\cite{ho2022classifier} with negative prompting. Given a positive text prompt $\bt_{\mathrm{pos}}$ for the desired sound effect and a negative prompt $\bt_{\mathrm{neg}}$ representing the original expected sound or an unconditional text embedding $\varnothing_t$, we compute a diffusion step via:
\begin{equation}
\hat{\boldsymbol{\epsilon}} =  (\gamma + 1) \cdot \dm(\bz_t, \bv_q, \bt_{\mathrm{pos}}, t)  
- \gamma \cdot \dm(\bz_t, \varnothing_v, \bt_{\mathrm{neg}}, t) \text{,} 
\label{eq:text-cfg}
\end{equation} 
where $\gamma$ controls the guidance strength, and $\varnothing_v$ is the unconditional embedding for the input videos.

\paragraph{Video-guided Foley with audio-visual control}
Our model enables generating synchronized soundtracks guided by both audio and video inputs. 
Our model can apply the sound characteristics (\eg, rhythm and timbre) of reference audio from an SFX library to a silent video, synchronize the audio with visual events, and enable control over the generated output based on the reference audio-visual conditions.
We frame this task as the video-guided audio extension problem, where we prepend the conditional audio latent $\bz_c = \mathcal{E}_a(\ba_c)$ and optional video feature $\mathcal{E}_v(\bv_c)$ to the noisy audio latents $\bz_T$ along the sequence dimension and apply masked denoising to generate the missing sound:
\begin{equation}
\hat{\boldsymbol{\epsilon}} =  (\gamma + 1) \cdot \dm\left([\bz_c; \bz_t], [ \bv_c;\bv_q], t \right)  
- \gamma \cdot \dm \left([\varnothing_a; \bz_t], \varnothing_v, t \right) \text{,} 
\label{eq:audio-cfg}
\end{equation} 
where $\varnothing_a$ denotes the noisy latent that matches the size of the conditional latent $\bz_c$. $\bv_c$ and $\bv_q$ are encoded by $\mathcal{E}_v$. 

\paragraph{Video-guided Foley with quality control}
We enforce quality control by incorporating quality tags in the text, enabling the generation of clean full-band (48kHz) audio. During inference, we guide the model to produce samples that align with the high-quality audio distribution while steering away from low-quality audio using CFG with a negative prompt $\bt_{\mathrm{neg}}$; the prompt can be ``low quality'' or the unconditional text embedding $\varnothing_t$:
\begin{align}
\hat{\boldsymbol{\epsilon}} =  (\gamma + 1) & \cdot \dm(\bz_t, \bv_q, \bt_c + \text{``high quality''} , t) \notag  \\
- & \gamma \cdot \dm(\bz_t, \varnothing_v, \bt_{\mathrm{neg}}, t) \text{.} 
\label{eq:quality-cfg}
\end{align}

\section{Experiments}
\label{sec:exp}
In this section, we quantitatively and qualitatively evaluate our method for the tasks of video-to-audio generation and video-guided Foley generation with multimodal controls over text, reference audio, and video.

\subsection{Video-to-Audio Generation}
First, we evaluate the ability of our model in the video-to-audio generation task, \ie, reproducing the soundtracks for silent videos, through an automatic quantitative evaluation.

\paragraph{Experimental setup}
We evaluate this task using videos from the VGGSound test set~\cite{chen2020vggsound}. To ensure accurate audio-visual correspondence, we apply ImageBind~\cite{girdhar2023imagebind} to filter out test samples with a score below 0.3, yielding a final set of 8,702 videos, following \cite{viertola2024temporally,xu2024video}. For each video, we generate 8-second audio samples.
We compare our model against several video-to-audio baselines, including the autoregressive models Im2Wav~\cite{sheffer2023hear} and FoleyGen~\cite{mei2023foleygen}, latent diffusion models Diff-Foley~\cite{luo2024diff}, V2A-mapper~\cite{wang2024v2a} and FoleyCrafter~\cite{zhang2024foleycrafter}, Frieren~\cite{wang2024frieren} based on rectified flow matching, and VAB~\cite{su2024vision} that apply MaskGIT~\cite{chang2022maskgit} framework. We also report the performance of the reconstructed audio with our DAC-VAE on the VGGSound test set as an oracle baseline. We trim the generated audio from the baselines to 8 seconds for a fair evaluation. Our model approaches the video-to-audio generation task as video-text-to-audio (VT2A) generation, using the VGGSound category name as the text input. During the inference, we use the ``low quality'' tag for our model's generation to stay within the same data distribution of VGGSound for a fair comparison and a guidance scale of 3.0 for diffusion sampling.

\paragraph{Evaluation metrics} 
Following prior work~\cite{mei2023foleygen,wang2024frieren,viertola2024temporally}, we evaluated model performance in terms of audio quality and cross-modal alignment. We evaluated audio quality using Fréchet Audio Distance (FAD)~\cite{kilgour2018fr} with VGGish~\cite{hershey2017cnn} on 16kHz audio, which measures the distribution distance between generated and reference audio. For reference sets, we used the VGGSound test set as regular references and Adobe SFX Audition\footnote{\url{https://www.adobe.com/products/audition/offers/adobeauditiondlcsfx.html}} (a professional audio library that differs from VGGSound) as high-quality references, denoting these metrics as FAD@VGG and FAD@AUD, respectively. We also use Kullback–Leibler Divergence (KLD)~\cite{kullback1951information} to measure the probability distributions of class predictions by the PaSST~\cite{koutini2021efficient} model between the ground-truth and generated samples.
To assess cross-modal alignment, we use ImageBind~\cite{girdhar2023imagebind} to measure the semantic correspondence between the generated audio and the input video. We also compute the CLAP score~\cite{wu2023large} to evaluate the similarity between category labels and generated audio. To measure the cross-modal (temporal) alignment between generated audio and input videos, we apply Synchformer~\cite{iashin2024synchformer} to estimate the weighted temporal offset in seconds, following \cite{viertola2024temporally}. The model classifies the offset from -2.0s and 2.0s (with a 0.2s resolution). The final \textit{AV-Sync} metric is the average of the absolute offsets across all examples.

\paragraph{Quantitative results}
We show our quantitative results in \cref{tab:video2audio} and \cref{fig:video2audio} and demonstrate that our method outperforms all the other methods in multiple metrics, including \textit{AV-Sync}, CLAP score, FAD@AUD, and KLD. This highlights the strong overall performance of our model against baselines. Notably, our synchronization score is comparable to the DAC-VAE reconstructed results, indicating that we successfully generated synced audio for silent input videos. Additionally, our method achieves the second-best performance on the ImageBind score, reflecting strong cross-modal semantic alignment. We outperform the oracle baseline (DAC-VAE) on the CLAP score indicating that our generated clips are more semantically aligned to corresponding sound effects.  
The performance of DAC-VAE on FAD@VGG indicates that the VAE influences the generated data distribution, which in turn affects the FAD evaluation.

\subsection{Video-guided Foley with Text Control}
We conduct experiments to evaluate our model’s capability for video-guided Foley generation with text controls, focusing on synchronization and diversity of semantics.

\paragraph{Experimental setup}
To quantitatively evaluate the semantic control of the model with text, we sampled videos of 10 categories from the VGGSound-sync~\cite{chen2021audio} dataset as a test set with the ImageBind~\cite{girdhar2023imagebind} filtering strategy described above. We individually provided the other 9 category names for each video as target text prompts and tasked the models with generating audio based on the given text and video. We generated 4 audio tracks for each pair. We compared our model with FoleyCrafter~\cite{zhang2024foleycrafter}, which also supports text-based control. We also modify FoleyCrafter to disable its semantic adapter to cut the semantic signal from input videos, using this model solely as a video-onset ControlNet on the text-to-audio generation model.
We evaluated four variants of our approach: 1) \textbf{w/ NegP}: using target prompts as positive prompts and the ground truth category as negative prompts for classifier-free guidance; 2) \textbf{w/o NegP}: no negative prompts are used; 3) \textbf{True category}: regular video-text-to-audio generation with true category as positive prompts (without negative prompts) to generate expected Foley sound for videos, representing the best synchronization performance that our models could achieve; 4) \textbf{T2A}: text-to-audio generation with the given target individual prompts, providing the upper bound for CLAP metrics. 

\paragraph{Evaluation metrics} 
To evaluate semantic alignment between target prompts and generated audio, we use two CLAP-based~\cite{wu2023large} metrics. First, we calculate the CLAP score, defined as the average cosine similarity between the CLAP embeddings of each text prompt and generated audio pair. Additionally, we use the CLAP model as a classifier to compare target category scores against original categories (from videos), reporting binary classification accuracy. We also report temporal performance using the aforementioned \textit{AV-Sync} metric. Although generated audio-visual examples fall outside the synchronization model’s training distribution, we found that it returns reliable scores.

\begin{table}
\caption{\textbf{Evaluation on the Foley generation with text controls.} {\it NegP} denotes negative prompting. We also include two oracle baselines: one using the true category as text prompts and the other omitting video during inference. The best results are in {\bf bold}. }
\label{tab:text-foley}
\newcommand\padd{\phantom{0}}
\setlength\tabcolsep{4pt}
\centering
\resizebox{1.0\columnwidth}{!}{%
\begin{tabular}{llccc}
\toprule
\multirow{2}{*}{Method} & \multirow{2}{*}{Variation} & \multicolumn{2}{c}{CLAP~$\uparrow$} & \multirow{2}{*}{AV-Sync~$\downarrow$} \\
\cmidrule(lr){3-4}
 &  & Score & Acc &  \\

\midrule
FoleyCrafter~\cite{zhang2024foleycrafter} & w/o NegP & \textbf{38.4} & 99.4 & 1.34 \\
 {\small(w/o semantic adapter)} & w/ NegP & 35.7 & \textbf{99.9}   & 1.36  \\
 \cdashlinelr{1-5}

\multirow{2}{*}{FoleyCrafter~\cite{zhang2024foleycrafter}} & w/o NegP & 31.0 & 79.2 & 1.29  \\
 & w/ NegP & 33.4 & 94.2   & 1.31  \\

\cdashlinelr{1-5}
\multirow{2}{*}{Ours} 
&  w/o NegP  & 31.4 & 85.5 & \textbf{0.81} \\
&  w/ NegP  & 30.9  &  93.2 & 0.93 \\
\midrule
\multirow{2}{*}{Ours -- oracle} 
&  True category & \padd 4.2 & \padd 1.8 & \textbf{0.77} \\
&  T2A  & \textbf{40.3} & \textbf{100} &  1.38 \\
\bottomrule
\end{tabular}%
}
\downvspacefig
\end{table}

\begin{figure*}[!t]
\upvspacefig
    \centering
    \includegraphics[width=\linewidth]{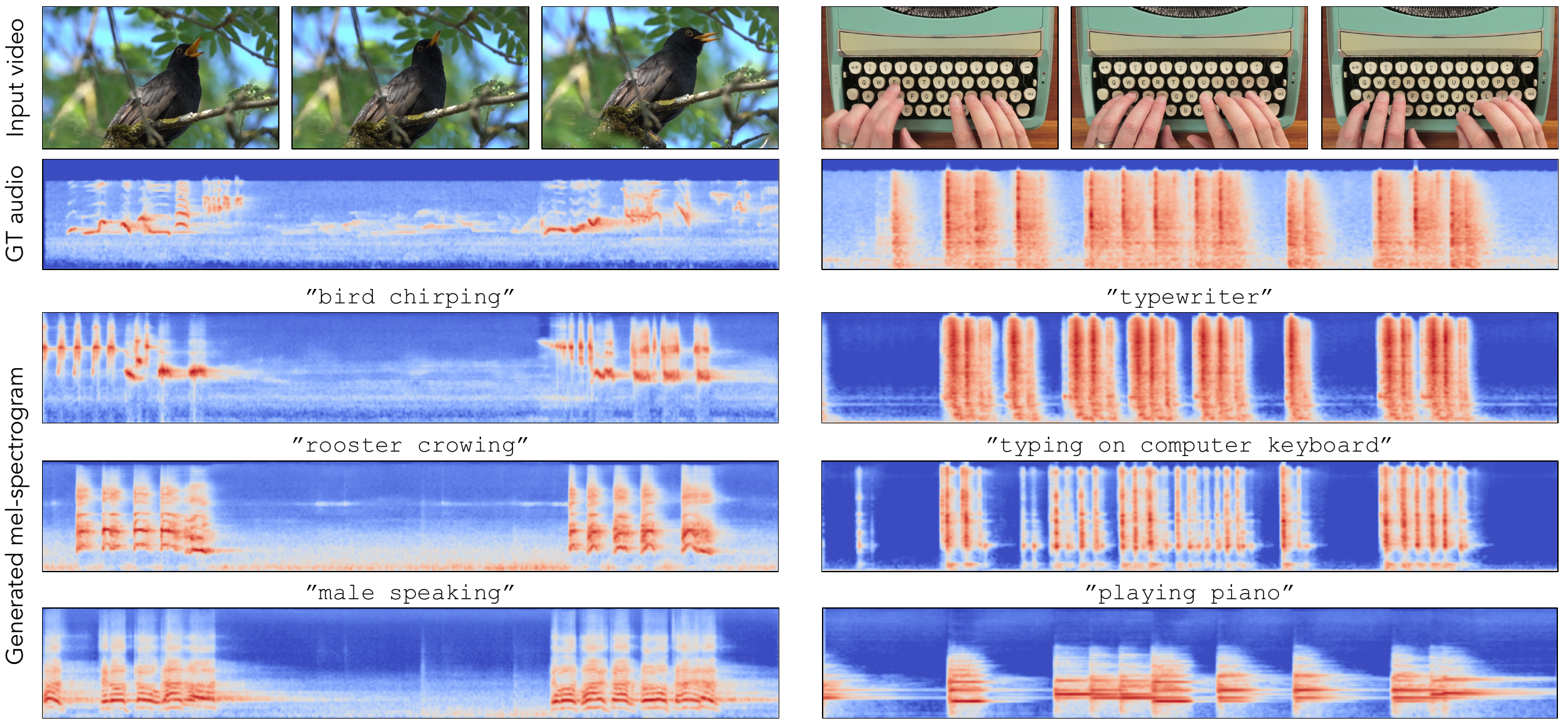}
    
    \caption{\textbf{Qualitative examples for Foley generation with text control.} We present generated results for two videos, each with three different text prompts, demonstrating our model's ability to produce synchronized soundtracks with varied semantics through text control. 
    {\bf Please refer to our \href{\projecturl}{website} for video results.}
    }
    \label{fig:text-control}
    \downvspacefig
\end{figure*}

\paragraph{Results}
We present the results in \cref{tab:text-foley}. Our method achieves the highest synchronization performance, comparable to the oracle model (Ours--true category) that generates soundtracks with original semantics. The CLAP metrics are also reasonable, indicating that our approach effectively generates synchronized audio for input videos while allowing semantic control through text. Although FoleyCrafter~\cite{zhang2024foleycrafter} performs well on CLAP-based metrics, its AV-Sync score is similar to text-to-audio generation, suggesting it struggles to generate synchronized audio in this task. Additionally, the results of the CLAP accuracy highlight that the negative prompting strategy helps steer the generation away from reproducing the original semantics of the audio, enhancing semantic control through classifier-free guidance while the sync score is slightly dropped. 
We also show some qualitative examples for text control in \cref{fig:text-control}, demonstrating that our model enables the disentangling of semantic information from the input video and maintains the temporal information. 

\paragraph{Human studies}
We conducted a human evaluation using two-alternative forced choice (2AFC) studies. We selected 10 high-quality videos from the VGGSound test set, ensuring diverse categories and clear temporal information. For each video, we used one prompt based on the original video's semantics and another that diverged from it. We compared our method with FoleyCrafter~\cite{zhang2024foleycrafter}.

In the user study, participants watch and listen to two video clips, each paired with an audio sample—one generated by our model and the other from the baseline. They were asked to select which audio (1) best matches the sound of ``{\texttt{\small audio prompt}}'', (2) is best synchronized with the video, (3) which audio sounds cleaner and more high definition, and (4) overall sounds best when considering the intended audio for ``{\texttt{\small audio prompt}}''. Additional details of the study can be found in  \supparxiv{the supplementary material}{\cref{appendix:user}}.

We demonstrate the user study results in \cref{tab:human-study}. As can be seen, our method outperforms the baseline in all the aspects. Human evaluators consistently rate our results as being higher in semantic alignment, audio-visual synchronization, and audio quality. It demonstrates our model’s capability in video-guided Foley generation with text control, as well as the quality of the generated audio.

\begin{table}
\caption{\textbf{Human Evaluation on the Foley generation with text control.} We show the win rates of our method against FoleyCrafter~\cite{zhang2024foleycrafter}. 95\% confidence intervals are reported in \textcolor{gray}{gray}. P-values are below $10^{-20}$. ($N=400$ with 20 participants) }
\label{tab:human-study}

\centering
\setlength\tabcolsep{3.5pt}
\resizebox{1\columnwidth}{!}{%
\begin{tabular}{lcccc}
\toprule
\multirow{2}{*}{Comparison}  & \multicolumn{4}{c}{Win rate~(\%)} \\
\cmidrule(lr){2-5}
&  Semantic & Sync. & Quality & Overall \\

\midrule

Ours vs FoleyCrafter & \textbf{85.8}\ci{3.4}  & \textbf{94.5}\ci{2.1}  & \textbf{86.5}\ci{3.4}  &  \textbf{90.2}\ci{2.9} \\
\bottomrule
\end{tabular}%
}
\downvspacefig
\end{table}

\begin{table}
\caption{\textbf{Evaluation on the Foley extension with different control signals}. $\mathcal{V}_q$ means input slient videos. $\mathcal{T}_c$,  $\mathcal{A}_c$ and  $\mathcal{V}_c$ denote text, audio and video conditional signals respectively. The best results are in {\bf bold}.}
\label{tab:foley-extension}
\newcommand\padd{\phantom{0}}
\centering
\resizebox{1.0\columnwidth}{!}{
\begin{tabular}{l cccc cc}
\toprule
\multirow{2}{*}{Eval set}  & \multicolumn{4}{c}{Conditions} & \multirow{2}{*}{CLAP~$\uparrow$} & \multirow{2}{*}{AV-Sync~$\downarrow$} \\
\cmidrule(lr){2-5}
 & $\mathcal{V}_q$  &  $\mathcal{T}_c$ & $\mathcal{A}_c$ & $\mathcal{V}_c$ &  \\
\midrule
\multirow{4}{*}{VGGSound} 
&  \cmark &  \cmark &   &    & 55.4  &  0.79 \\
&  \cmark &   & \cmark  &   &  59.6 &  0.78 \\
&  \cmark &   & \cmark  & \cmark  &  59.8 & 0.77\\
&  \cmark &  \cmark & \cmark  & \cmark  & \textbf{64.3} & \textbf{0.77} \\
\midrule
\multirow{3}{*}{Greatest-Hits} 
&  \cmark &  &   &    & 29.3 & 0.88      \\
&  \cmark &   & \cmark  &   & 73.8 &  0.94  \\
&  \cmark &   & \cmark  & \cmark  & \textbf{74.4}  & \textbf{0.87} \\
\bottomrule
\end{tabular}%
}
\downvspacefig
\end{table}

\begin{table*}[!t]
\upvspacefig
\caption{{\bf Quantitative evaluation on quality control and ablation study.} We evaluate our model with different inference settings, \ie, using various quality tags and excluding text input. We also ablate the impact of the subset fine-tuning strategy. {\it NegP} denotes negative prompting. The best results are in {\bf bold}. }
\label{tab:ablation}

\centering
\footnotesize
\renewcommand{\arraystretch}{1.0}
\newcommand\padd{\phantom{0}}

\resizebox{1\textwidth}{!}{

\begin{tabular}{clll ccc ccc}
\toprule

& Variation  & Inference  & Quality tag  & ImageBind~$\uparrow$ & CLAP~$\uparrow$ & AV-Sync~$\downarrow$  &  FAD@VGG~$\downarrow$ & FAD@AUD~$\downarrow$ & KLD~$\downarrow$   \\
\midrule
\multirow{4}{*}{Ours}
& Full w/o ft.  & VT2A & Low    &  27.3     &  33.8 & 0.81  & 3.00 & 4.39  & 1.47\\

& Full &   VT2A &  Low   &  \textbf{ 28.0 }  &  34.4 & 0.80 & \textbf{2.92} & 4.62  & \textbf{1.43}\\
& Full &  VT2A  & High  & 25.8 &  \textbf{34.9}  & 0.83 &  4.37   &  4.09 & 1.60 \\

&  Full & V2A~(w/o text)  & -    & 22.4  & 19.4 & \textbf{0.77}  & 4.78  & \textbf{3.44} &  2.59\\

\bottomrule
\end{tabular}

}
\downvspacefig
\end{table*}

\subsection{Video-guided Foley with Audio-Visual Control}
Our model emerges with the ability to generate video-guided Foley using reference audio-visual examples, effectively transferring sounds from conditional clips to generate synchronized soundtracks for silent videos. We evaluate this on the Foley extension task, where the first few seconds of a sounding video serve as the condition, and the model generates the remaining audio for the silent part of the video.

\paragraph{Experiment setup}
We evaluate this task on two datasets: VGGSound-Sync~\cite{chen2021audio} and Greatest-Hits~\cite{owens2016visually}. We sample 1,000 8-second video examples from VGGSound-Sync as in-domain data, and 800 8-second examples from Greatest-Hits as out-of-domain data. For each video, the first 3 seconds of both audio and video ($\ba_c, \bv_c$) serve as conditional inputs, while the model generates the remaining 5 seconds of audio $\hat{\ba}$ for the corresponding silent video segment $\bv_q$.
We evaluate the results using two metrics: (1) CLAP score to measure the cosine similarity between the ground-truth and generated audio, and (2) AV-Sync score to evaluate synchronization accuracy.
We generate four audio clips for each video and compare our model’s performance across different combinations of multimodal conditional inputs ($\bt_c$, $\ba_c$, $\bv_c$). Text inputs are omitted to evaluate the Greatest-Hits dataset, focusing solely on the audio and video conditions.

\paragraph{Results}
We present the results in \cref{tab:foley-extension}. Across both datasets, CLAP similarity improves substantially with the addition of audio conditions. When video conditions are added, we observe further improvement in the AV-Sync score, particularly for the Greatest-Hits dataset. This suggests that the model effectively leverages in-context learning to interpret information from conditional audio-visual inputs.

Additionally, our model demonstrates strong capability in Foley analogy tasks as shown in \cref{fig:teaser}. For instance, given a reference dog bark, the model can produce synchronized Foley audio for a different dog barking video, or generate drum sounds for a basketball dribble video with a reference drum audio sample. Please see our \href{\projecturl}{website} for examples.

\subsection{Video-guided Foley with Quality Control}
\begin{figure}[!t]
    \centering
    \includegraphics[width=\linewidth]{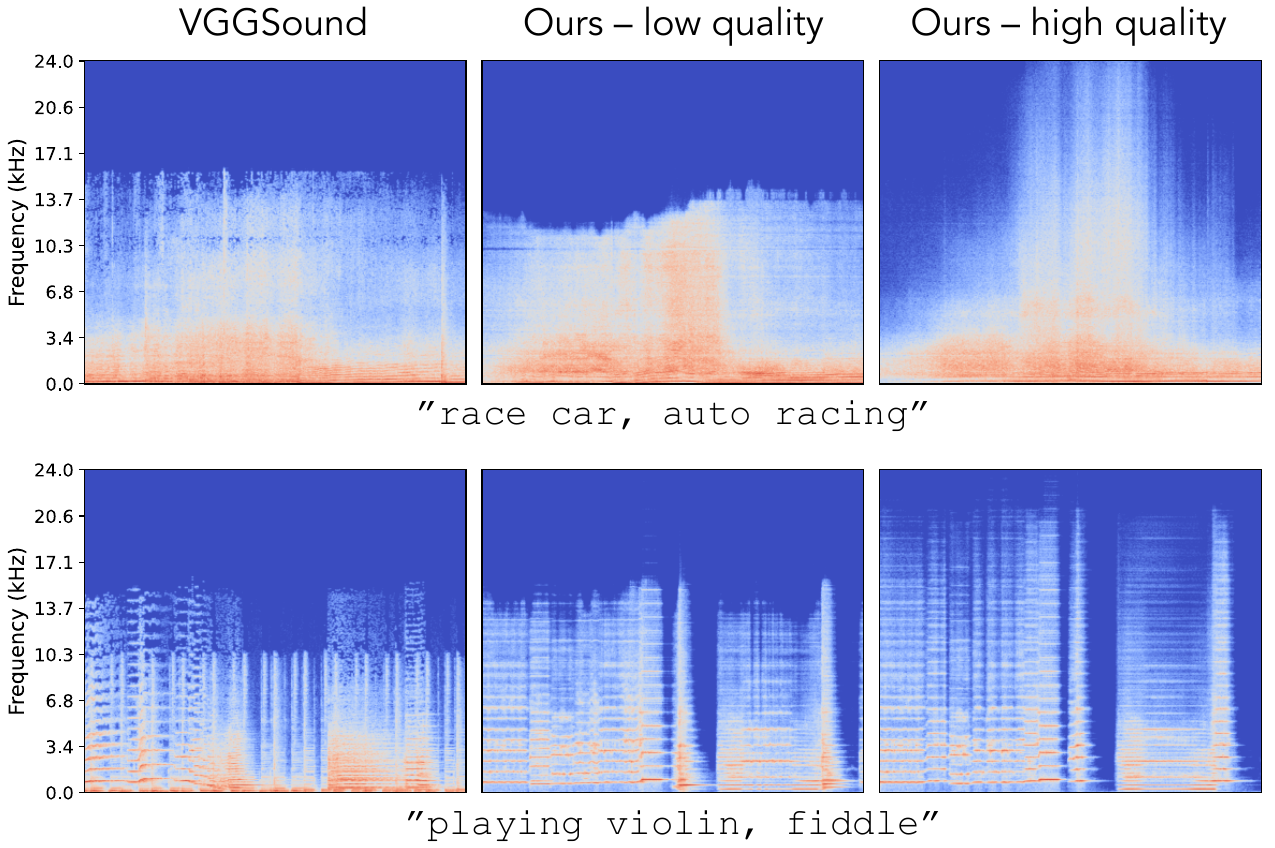}
    
    \caption{\textbf{Qualitative results of quality control.} We show that VGGSound audio has limited bandwidth and demonstrate our model generates full-band 48kHz audio with quality control.  
    } 
    \label{fig:quality-control}
    \downvspacefig
\end{figure}

During our experiments, we observed that audio from VGGSound videos downloaded via YouTube is often compressed to 32kHz or lower. When resampled to 48kHz, the high-frequency content above 32kHz is missing, as shown in \cref{fig:quality-control}. Consequently, when training on the VGGSound dataset, models inherently generate audio with a quality aligned to 32kHz, reflecting the dataset’s distribution.

Our model incorporates high-quality text-audio data at 48kHz during training to address this limitation, paired with quality tags. This enables the generation of full-band, 48kHz audio for video-guided Foley generation. During inference, we use the ``high quality'' tag to guide the model toward generating audio that follows the distribution of the high-quality text-audio dataset, ensuring 48kHz output.

We provide examples in \cref{fig:quality-control}, demonstrating the model’s ability to control and improve audio quality. Additionally, quantitative results in \cref{tab:ablation} support this, where we observe that with the ``high quality'' tag, our model achieves better performance on the FAD@AUD metric, while performance on FAD@VGG decreases. This suggests the model effectively generates audio that aligns more closely with high-quality sound effect distributions.

\subsection{Ablation}
\paragraph{Subset fine-tuning}
As discussed in \cref{sec:model-recipe}, the VGGSound dataset includes noisy, low-quality samples where audio and video often lack alignment. To enhance model performance, we further fine-tune our pretrained model on a subset of VGGSound containing better audio-visual correspondence. We evaluate our pretrained model performance and compare it with the fine-tuned model. We present results in \cref{tab:ablation}, demonstrating that fine-tuning significantly improves cross-modal alignment metrics.

\paragraph{No-text inference}
We also perform an ablation study where we remove text conditioning and evaluate our model directly on video-to-audio generation. We report the results in \cref{tab:ablation}. We observe that while performance on the semantics-related metrics drops significantly, synchronization metrics remain high, indicating that our model relies on text to drive semantic alignment but can retain temporal consistency even in the absence of any conditioning.

\section{Conclusion}
\label{sec:conclusion}

In this paper, we present \textsc{MultiFoley}, a Foley system designed for video-guided Foley generation using multimodal inputs, including text, audio, and video.  
We evaluate our model on standard video-to-audio generation tasks, providing quantitative evidence of its effectiveness. 
We explore the model’s control capabilities with different conditional inputs through both quantitative and qualitative experiments, illustrating a range of Foley applications achievable with our approach. 
Our work is a step toward the broader goal of ``user-in-the-loop'' sound design.
By providing easy-to-use, multimodal controls, we aim to help users create customized, synchronized high-quality audio.

\paragraph{Limitations and broader impacts}
Our model is currently trained on a small-scale, in-the-wild audio-visual dataset, VGGSound, which constrains its capabilities. We believe a larger, high-quality Foley dataset would significantly enhance our model’s performance and broaden its applicability.
Our model currently struggles with handling multiple sound events alongside text, often leading to confusion about the timing of each event.
Our method can also create realistic but potentially misleading media. Responsible use is essential to prevent misuse in situations where authenticity matters.

\paragraph{Acknowledgements}

We thank Sonal Kumar, Hugo Flores García, Xiulong Liu, Yongqi Wang, and Yiming Zhang for their valuable help and discussion, and Adolfo Hernandez Santisteban for informing the project with his sound design experience.
This work was supported in part by an NSF CAREER Award \#2339071.

{
    \small
    \bibliographystyle{ieeenat_fullname}
    \bibliography{main}
}

\clearpage

\appendix
\supparxiv{
\setcounter{page}{1}
\maketitlesupplementary
}{}

\renewcommand{\thesection}{A.\arabic{section}}
\setcounter{section}{0}

\section{Implementation Details}
\label{appendix:implement}

\paragraph{DAC-VAE} 
We implemented and trained a modified version of the Descript Audio Codec (DAC)~\cite{kumar2024high} using a variational autoencoder (VAE)~\cite{kingma2013auto}. In this approach, we replaced the residual vector quantizer (RVQ) with a VAE objective to encode continuous latents, enabling diffusion models to operate on continuous representations instead of discrete tokens. Our DAC-VAE was trained on audio waveforms at various sampling rates, allowing us to encode a 48kHz waveform into latents at a 40Hz sampling rate, with a feature dimension of 64. We train our DAC-VAE model on a variety of proprietary and licensed data spanning speech, music, and everyday sounds.

\paragraph{DiT architecture} 
Our DiT model has 12 layers, each with a hidden dimension of 1024, 8 attention heads, and an FFN (Feed-Forward Network) dimension of 3072, totaling 332M parameters. For the audio latents, we use an MLP (Multi-Layer Perceptron) to project them into 512-dimensional features. A separate MLP maps encoded visual features to 512 dimensions, followed by nearest-neighbor interpolation to upsample them fivefold (from 8Hz to 40Hz). Finally, we concatenate the audio and video features along the channel dimension to form 1024-dimensional inputs, which are then fed into the transformer.

Similar to VampNet~\cite{garcia2023vampnet}, we use two learnable embeddings to differentiate between conditional input audio latents and noisy latents to be denoised, based on the conditional mask. We then sum the corresponding mask embeddings to the audio latents. During the inference, we create a conditional mask to achieve audio-conditioned generation.

\paragraph{Training details} 
We use the AdamW optimizer~\cite{kingma2015adam,loshchilov2017decoupled} with a learning rate of $10^{-4}$ and apply a cosine decay schedule. Training begins with a linear warm-up phase for the first 4K iterations, followed by 599.6K iterations. We train our model with Exponential Moving Average (EMA)~\cite{morales2024exponential} with EMA decay of 0.99.
Throughout the training, we randomly sample from combined datasets where 60\% of training examples are from VGGSound and 40\% from HQ-SFX. Within VGGSound samples, 60\% are dedicated to video-text-tag-to-audio generation, the rest 40\% are evenly distributed across different dropout variants (\ie, video+tag, video+text,  video-only, text+tag, text-only, tag-only, unconditional). For HQ-SFX samples, 60\% are allocated to text-tag-to-audio generation, with the remaining cases divided as follows: 10\% for text-only, 15\% for tag-only, and 15\% for unconditional audio generation.

\section{Additional Experiments}
\label{appendix:exp}

\paragraph{Guidance scale ablation}
We also examine the effect of the classifier-free guidance (CFG) scale, as shown in \cref{tab:guidance}. The model shows similar performance with guidance weights between 3.0 and 7.0. On the FAD metrics, a higher guidance scale improves FAD@AUD but worsens FAD@VGG, suggesting that the model generates examples that align more with high-quality distributions. We use guidance scales of 3.0 and 5.0 for experiments in the main paper.

\section{Human Studies}
\label{appendix:user}

\paragraph{Videos and prompts}
We handpicked 10 high-quality videos from the VGGSound test set, choosing examples that span a variety of categories and contain clear, easily perceivable temporal actions. We crafted two text prompts for each video: one matching the original category and another for a different target category, shown in \cref{tab:human_prompt}. We then generated four 8-second samples for each video and randomly selected one for the final evaluation in the survey. For our model's generation, we use the ``high quality'' tag for inference.

\begin{table}[!h]
  \caption{{\bf Audio prompts for the user studies.} We note that the prompts are paired for the same video. }
  \label{tab:human_prompt}
  \setlength{\tabcolsep}{20pt}
  \renewcommand{\arraystretch}{1.0}
  \newcommand\padd{\phantom{0}}
  \centering
  \resizebox{\columnwidth}{!}{
  \begin{tabular}{ll}
    \toprule
    Original prompt  & ReFoley prompt \\
    \midrule 
    playing cello & playing erhu \\
    bird chirping & rooster crowing \\
    dog barking & playing drum\\
    typewriter &  playing piano \\
    gunshot & snare drum playing \\
    chopping wood & kick drum playing \\
    lion roaring & cat meowing \\
    squeezing toys & cracking bones \\
    playing trumpet & playing saxophone \\
    playing golf & explosion \\
    \bottomrule
  \end{tabular}
  }
\end{table}

\begin{table*}[!t]
\caption{{\bf Ablation study for classifier-free guidance scale on video-to-audio generation.} The best results are in {\bf bold}. }
\label{tab:guidance}

\centering
\footnotesize
\setlength\tabcolsep{8pt}
\renewcommand{\arraystretch}{1.0}
\newcommand\padd{\phantom{0}}

\resizebox{1\textwidth}{!}{

\begin{tabular}{cl ccc ccc}
\toprule

& Variation  & ImageBind~$\uparrow$ & CLAP~$\uparrow$ & AV-Sync~$\downarrow$  &  FAD@VGG~$\downarrow$ & FAD@AUD~$\downarrow$ & KLD~$\downarrow$    \\
\midrule
\multirow{5}{*}{Ours} 
& $\gamma = 1.0$   &   26.4    & 32.4  & 0.90   & 3.16    &   4.27  & 1.49  \\
& $\gamma = 3.0$  & 28.0   &  34.4 & 0.80  & \textbf{2.92} & 4.62  & \textbf{1.43}  \\
& $\gamma = 4.0$  & \textbf{28.1}   &  34.7 & 0.77  & 3.05 & 4.59  & \textbf{1.43}  \\
& $\gamma = 5.0$ &  28.0    & \textbf{34.8}  &  0.77  & 3.27    &  4.48   &   \textbf{1.43}   \\ 
& $\gamma = 7.0$ &   27.5    &  34.6 &   \textbf{0.75} &  3.84   &   \textbf{4.21}  &   1.44  \\

\bottomrule
\end{tabular}

}
\end{table*}

\begin{figure*}[t]
\centering

\includegraphics[width=1.0\linewidth]{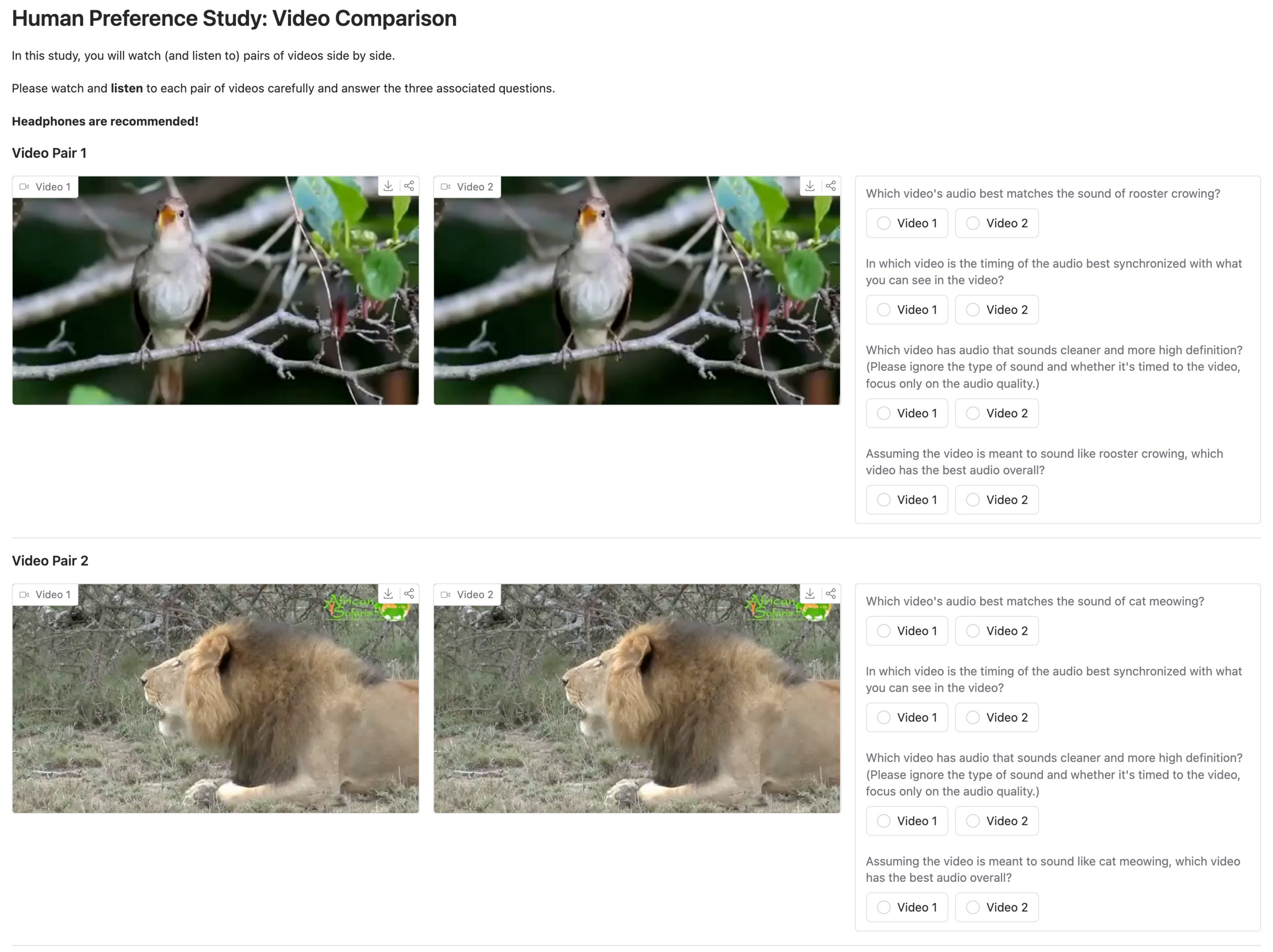}

\caption{ {\bf Screenshot of Foley user study.}  We show the screenshot from our user study survey. We show the instructions and the first two video pair examples and associated questions. }

\label{fig:human-study-screenshot}
\end{figure*}
\paragraph{User study survey}
In the survey, participants watched and listened to 20 pairs of videos comparing our method with FoleyCrafter~\cite{zhang2024foleycrafter}. We performed a forced-choice experiment where we randomized the left-right presentation order of the video pairs. For each video pair, participants were asked to respond to four questions:

\begin{enumerate}
    \item Which video's audio best matches the sound of \{{\tt \small audio prompt}\}?
    \item In which video is the timing of the audio best synchronized with what you can see in the video?
    \item Which video has audio that sounds cleaner and more high definition? (Please ignore the type of sound and whether it's timed to the video, focus only on the audio quality.)
    \item Assuming the video is meant to sound like \{{\tt \small audio prompt}\}, which video has the best audio overall?
\end{enumerate}

The first question evaluates the semantic alignment between the generated audio and the target audio prompt, ensuring that the sound matches the expected content. 
The second question evaluates the temporal alignment between the audio and video, focusing on how well the sound synchronizes with visual cues. The third question ignores content and timing to focus specifically on audio quality, examining aspects such as fidelity and production standards. Finally, the last question offers a holistic evaluation, determining which model produces the most effective overall audio. We show a screenshot of our user study survey including the instruction block, the first two video pairs, and associated questions in \cref{fig:human-study-screenshot}.

\end{document}